\documentclass{article} 

  \usepackage{iclr2023_modified_style,times}
 
\usepackage{amsmath,amsfonts,bm}

\def\eqref#1{equation~\ref{#1}}

\def\1{\bm{1}}

\DeclareMathAlphabet{\mathsfit}{\encodingdefault}{\sfdefault}{m}{sl}
\SetMathAlphabet{\mathsfit}{bold}{\encodingdefault}{\sfdefault}{bx}{n}

\DeclareMathOperator*{\argmin}{arg\,min}

\usepackage{hyperref}
\usepackage{url}

\newcommand{\norm}[1]{\left\|#1\right\|}

\usepackage[vlined,ruled]{algorithm2e}

\usepackage[utf8]{inputenc} 
\usepackage[T1]{fontenc}    
\usepackage{hyperref}       
\usepackage{url}            
\usepackage{booktabs}       
\usepackage{amsfonts}       
\usepackage{nicefrac}       
\usepackage{microtype}      
\usepackage{thm-restate}

\usepackage{microtype}
\usepackage{graphicx}
\usepackage{subfigure}
\usepackage{changes}

\usepackage{amsmath}
\usepackage{multirow}
\usepackage{hhline}
\usepackage{wrapfig}
\usepackage{floatrow}
\usepackage{caption}
\usepackage{enumitem}
\usepackage[export]{adjustbox}
\usepackage{amsmath,mathtools}

\usepackage{algorithmic}

\makeatletter
\newcommand{\raisemath}[1]{\mathpalette{\raisem@th{#1}}}
\newcommand{\raisem@th}[3]{\raisebox{#1}{$#2#3$}}
\makeatother

\DeclarePairedDelimiter\abs{\lvert}{\rvert}%

\newcommand{\uglad}{{\texttt{uGLAD}~}}
\newcommand{\ugladns}{{\texttt{uGLAD}}}
\newcommand{\glad}{{\texttt{GLAD}~}}
\newcommand{\ngr}{{\texttt{NGR}~}}
\newcommand{\ngrns}{{\texttt{NGR}}}
\newcommand{\ngrs}{{\texttt{NGRs}~}}
\newcommand{\ngrsns}{{\texttt{NGRs}}}
\newcommand{\ngm}{{\texttt{NGM}~}}
\newcommand{\ngms}{{\texttt{NGMs}~}}
\newcommand{\ngmsns}{{\texttt{NGMs}}}
\newcommand{\gcpn}{{\texttt{GcPn}~}}

\title{Neural Graph Revealers}

\author{
  ~~Harsh Shrivastava~~~~
  Urszula Chajewska
  \hspace{0mm}\\
  \hspace{-3mm}
  \begin{tabular}{c}
      $\prescript{}{}{\text{ }~~\quad\text{Microsoft Research, Redmond, USA}}$
  \end{tabular}
}
\iclrfinalcopy 
\begin{document}

\maketitle
\begin{abstract}
Sparse graph recovery methods work well where the data follows their assumptions but often they are not designed for doing downstream probabilistic queries. This limits their adoption to only identifying connections among the input variables. On the other hand, the Probabilistic Graphical Models (PGMs) assume an underlying base graph between variables and learns a distribution over them. PGM design choices are carefully made such that the inference \& sampling algorithms are efficient. This brings in certain restrictions and often simplifying assumptions. In this work, we propose Neural Graph Revealers (\ngrsns), that are an attempt to efficiently merge the sparse graph recovery methods with PGMs into a single flow. The problem setting consists of an input data $X$ with $D$ features and $M$ samples and the task is to recover a sparse graph showing connection between the features and learn a probability distribution over the $D$ at the same time. \ngrs view the neural networks as a `glass box' or more specifically as a multitask learning framework. We introduce `Graph-constrained path norm' that \ngrs leverage to learn a graphical model that captures complex non-linear functional dependencies between the features in the form of an undirected sparse graph. Furthermore, \ngrs can handle multimodal inputs like images, text, categorical data, embeddings etc. which is not straightforward to incorporate in the existing methods. We show experimental results of doing sparse graph recovery and probabilistic inference on data from Gaussian graphical models and a multimodal infant mortality dataset by Centers for Disease Control and Prevention.
\\

\textit{Keywords}: Sparse Graph Recovery, Deep Learning, Probabilistic Graphical Models\\
\textit{Software}: {\small\url{https://github.com/Harshs27/neural-graph-revealers}}
\end{abstract}

\section{Introduction}


Sparse graph recovery is an important tool to gain insights from data and is a widely researched topic~\cite{heckerman1995learning,bnlearn,shrivastava2022methods}. Given an input data $X$ with $M$ samples and $D$ features, the graph recovery algorithms discover the feature dependencies in the form of a sparse graph, $S_G\in\mathbb{R}^{D\times D}$, where $S_G$ is the adjacency matrix. Such graphs are useful for analyzing data from various domains. For instance, obtaining gene regulatory networks from single-cell RNA sequencing data or microarray expression data~\cite{haury2012tigress,moerman2019grnboost2,shrivastava2020grnular,shrivastava2022grnular}. Similarly, in finance domain, a graph showing correspondence between various companies can be obtained using the stock market data and navigation purposes~\cite{hallac2017network}. Other interesting applications consists of studying brain connectivity patterns in autistic patients~\cite{pu2021learning}, increasing methane yield in anaerobic digestion process~\cite{shrivastava2022uglad,shrivastava2022a} and gaining insights from the infant-mortality data from CDC in the US~\cite{shrivastava2022neural}. These sparse graphs capture functional dependencies and can be directed, undirected or in some cases have mixed-edge types. 

The space of graph recovery algorithms is quite substantial (see Fig.~\ref{fig:graph-recovery-undirected}), so to narrow down the scope of exploration, we only consider the problem of recovering undirected graphs in this work. The desired algorithm for recovering sparse graphs should have the following properties, (I) A rich functional representation to capture complex functional dependencies among features, (II) Handle diverse data types within the same framework, (III) Enforce sparsity, (IV) Preferably unsupervised as acquiring ground truth data can be expensive and at times a bit tricky, (V) Efficient \& scalable to handle large number of features.  

It is a difficult to design methods that can achieve a good balance between the desiderata mentioned. Many existing approaches make a simplifying assumption about the distribution in order to achieve sparsity~\cite{friedman2008sparse,van2010inferring,haury2012tigress,belilovsky2017learning,moerman2019grnboost2}. Some use the deep unfolding technique to use the existing optimization algorithm as an inductive bias for designing deep learning architectures~\cite{shrivastava2019cooperative,shrivastava2020using}. These deep unfolding based methods can achieve scalability and sparsity at the cost of making assumption about the underlying joint probability distribution~\cite{shrivastava2019glad,pu2021learning,shrivastava2022uglad} but require supervision for learning. The related methods section will do a walk-through of the associated algorithms and their formulations. 

In this work, we present an efficient algorithm, called Neural Graph Revealers (\texttt{NGRs}), that aspires to achieve the optimum balance between the listed desiderata. \ngrs can model highly non-linear and complex functional dependencies between the features by leveraging the expressive power of neural networks (NNs). It is a regression based approach that takes the $D$ features as input and maps them to the same features as the output. In order to achieve sparsity without compromising on the function representation capacity, \ngrs build up on the neat idea of viewing the neural networks as a `glass box'. Specifically, the NNs can be considered as a multitask learning framework between the input and output units. The paths between the input and output units are used to capture feature dependencies and thus restricting these paths (eg. using path-norms) can enforce desired sparsity in an unsupervised manner, refer Fig.~\ref{fig:ngr-architecture}. 
Thus, \ngrs recover undirected graphs that reveal the functional dependencies between the features. Furthermore, the design of \ngrs adhere to a special type of probabilistic graphical models known as Neural Graphical Models~\cite{shrivastava2022neural}. Thus, after learning the \ngr parameters, one can also query them for probabilistic reasoning, inference and sampling tasks.

Listing down the key features and contributions of this work.
\begin{itemize}[leftmargin=*,nolistsep]
    \item Novel use of neural networks as a multi-task learning framework to model functional dependencies that enables richer \& complex representation as compared to the state-of-the-art methods.
    \item Can incorporate multi-modal feature types like images, text, categorical data or generic embeddings. 
    \item Training is unsupervised which facilitates wider applicability and adoption to newer domains.
    \item Efficient and scalable approach that can handle large number of features.
    \item Once learned, the \ngr architecture becomes an instance of Neural Graphical Models and can be used for downstream probabilistic reasoning tasks.
\end{itemize}

\section{Related Methods}

\begin{figure}
\centering 
\includegraphics[width=135mm]{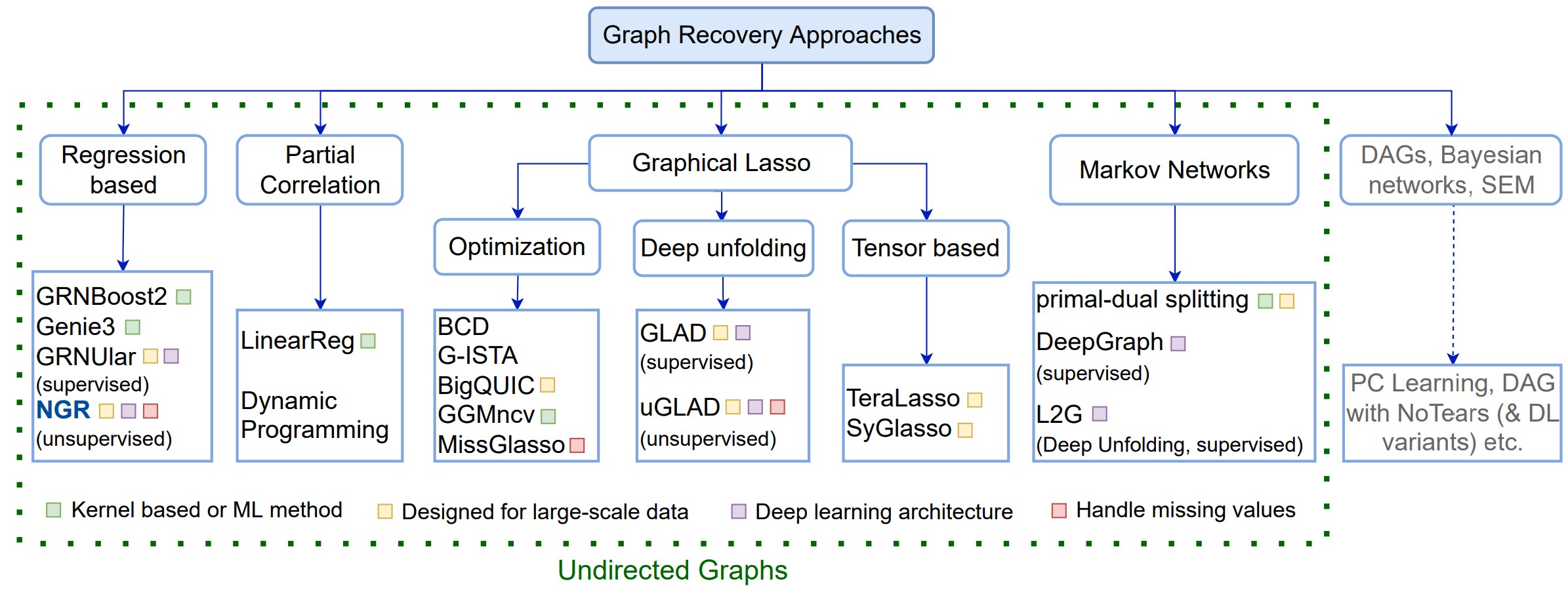}
\caption{\small \textbf{Graph Recovery approaches.} Methods designed to recover undirected graphs are categorized. Neural Graph Revealers (\texttt{NGRs}) lie under the regression based algorithms. The algorithms (leaf nodes) listed here are representative of the sub-category and the list is not exhaustive.
}
\label{fig:graph-recovery-undirected}
\end{figure}

Fig.~\ref{fig:graph-recovery-undirected} is an attempt to categorize different methods to recover graphs. Broadly saying, we can divide them as approaches recovering directed graphs and the ones recovering undirected or mixed-edge graphs. In this work, we primarily focus on methods developed for undirected graphs. 
Directed graphs have different semantics, with more complex dependence ralationships encoded by edges.  They are sometimes used to determine causality. Though an important research area, it is still theoretically at an early stage and requires very strong assumptions~\cite{pearl2018book}. Bayesian network graphs  can be converted to their undirected versions by the process of moralization~\cite{koller2009probabilistic}. We focus on modeling pairwise dependencies and not causation in this work, so undirected graphs will suffice.
For the purpose of categorizing different methods that recover undirected graphs, we break-down methods as the ones based on solving the graphical lasso objective also known as Conditional Independence (CI) graphs, the ones directly evaluating the partial correlations versus the ones based on the regression formulation. The remaining methods lie in the generic category of recovering Markov networks. 

\textit{Regression based approaches.} Consider the input data $X\in\mathbb{R}^{M\times D}$ that has $D$ features and a total of $M$ samples. For each of the $D$ features, the regression based formulation fits a regression function with respect to all the other features, $X_d=f_d(X_{\{D\}\backslash d})+\epsilon$, where $\epsilon$ is additive noise term. After fitting the regression, based on the choice of functions $f_d's$, the algorithms determine the dependency of the features. These approaches have been very successful for the task of recovering Gene Regulatory Networks. For instance, GENIE3~\cite{van2010inferring} modeled each $f_d$ to be random forest model while TIGRESS~\citep{haury2012tigress} chose $f_d$ as a linear function and  GRNBoost2~\citep{moerman2019grnboost2} combined random forests with gradient boosting technique to achieve superior performance among others~\cite{yu2002using,margolin2006aracne,van2020scalable,aluru2021engrain}. Then, neural network based representation like GRNUlar~\citep{shrivastava2020grnular,shrivastava2022grnular} which also utilized the idea of using NNs as a multitask learning setup were developed. This method is architecturally quite close to our formulation, although the major difference with \ngrs is that GRNUlar needs supervision for training. The proposed \ngr method can be categorized as a regression based approach. Unlike most other approaches in this group, it can be extended to model features that are not real-valued.

\textit{Partial Correlation evaluation.} Methods belonging to this category aim to directly calculate the partial correlations between all pairs of features to determine the Conditional Independence (CI) graph between the features. Algorithms based on fitting linear regression or using dynamic programming to evaluate the partial correlation formula directly have been developed~\cite{shrivastava2022methods}.  

\textit{Graphical Lasso formulation.} Doing sparse graph recovery based on the graphical lasso formulation and its variants have been extensively studied and a recent survey~\cite{shrivastava2022methods} provides a good primer into these approaches.  The traditional optimization based algorithms like BCD~\cite{banerjee2008model}, G-ISTA~\cite{rolfs2012iterative}, BigQUIC~\cite{hsieh2014quic}, GGMncv~\cite{williams2020beyond}, MissGlasso~\cite{stadler2012missing} and many others have been designed for a range of requirements like scaling to large number of features, handling missing values, including non-convex penalties etc. The TeraLasso~\cite{greenewald2019tensor} and the Sylvester Graphical Lasso or SyGlasso model~\cite{wang2020sylvester} are tensor based approaches and have recently garnered more interest. The methods in this category assume a multivariate Gaussian as the underlying distribution which might not be suitable for certain problems. The deep unfolding based approaches have shown to capture tail distributions and demonstrate better sample complexity results, \glad\cite{shrivastava2019glad} \& its unsupervised version \uglad\cite{shrivastava2022a}, can perform competitively with the existing approaches.

\textit{Markov Networks.} Probabilistic Graphical Models defined over undirected graphs are known as Markov Networks (MNs). Structure learning of MNs has been discussed extensively in~\cite{lee2006efficient,koller2009probabilistic,gogate2010learning,shrivastava2022methods}. We would like to point out some of the recent deep learning based methods that can capture rich functional dependencies like DeepGraph~\cite{belilovsky2017learning}, which uses convolutional neural networks to learn graph structure and L2G~\cite{pu2021learning} that uses the deep unfolding technique to convert optimization algorithm templates into deep architecture. Though these methods are interesting and better than their predecessors, they require supervision for training which hinders their wider adoption.   

\textit{Directed graph recovery approaches.} Bayesian Networks and causal models like Structural Equation models are represented using Directed Acyclic graphs (DAGs)~\citep{koller2009probabilistic,pearl2018book}. The structure learning problem of Bayesian Networks is NP-complete~\cite{chickering1996learning}, so often heuristic-based algorithms are proposed~\cite{singh1993algorithm}. Since our work focuses on recovering undirected graphs, we use this opportunity to draw out some similarities between \ngrs and the methods developed for DAG recovery. In a recent work, DAG with NoTears~\cite{zheng2018dags}, the authors converted the combinatorial optimization problem into a continuous one and provided an associated optimization algorithm to recover sparse DAGs. This eventually spawned research that provided improvements and also capture complex functional dependencies by introducing deep learning variants~\citep{zhang2019d,yu2019dag}. One such follow up work is by~\cite{zheng2020learning}, which we found to be the close to our method in terms of function representation capacity, specifically the MLP version of the nonparametric DAG learning. It uses a separate neural network for modeling each functional dependency of the graph. Although this leads to performance improvements, it also results in having significantly large number of learnable parameters as compared to \ngrs that model all the complex dependencies using a single NN by leveraging the multitask learning framework insight. 

Most of these methods were developed for real or numerical input data $X\in\mathbb{R}^{M\times D}$ and it is not straightforward to extend them for multi-modal feature types like images, text, categorical data or embeddings. \ngrs on the other hand provide a flexible approach to model such multi-modal inputs. 

\section{Neural Graph Revealers}

 We consider the basic setting where we are given the input data $X$
 with D features and M samples. The task is to recover a sparse graph represented by its adjacency matrix form $S_G\in\mathbb{R}^{D\times D}$. In the recovered undirected graph obtained by any regression based approach, each feature or graph node is a function of its immediate (one-hop) neighbors, as shown in the Fig.~\ref{fig:ngr-architecture}(right). In this section, we describe our proposed method Neural Graph Revealers (\texttt{NGRs}) along with its potential extensions.

\subsection{Representation}

\begin{figure}
\centering 
\includegraphics[width=135mm]{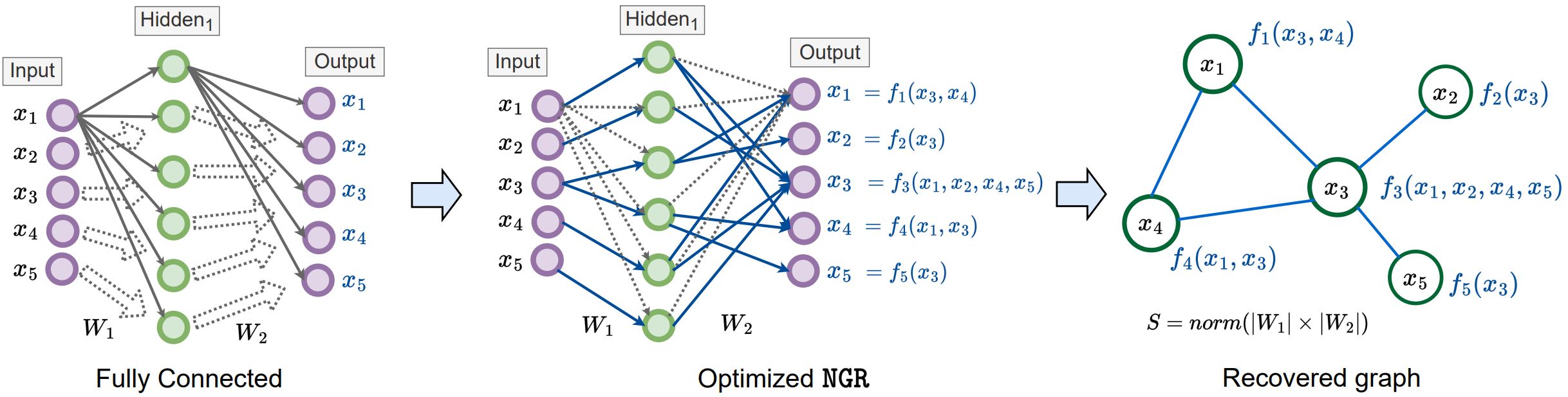}
\caption{\small \textit{Workflow of \ngrs.} (left) We start with a fully connected Neural Network (MLP here) where both the input and output are the given features $x_i's$. Viewing NN as a multitask learning framework indicates that the output features are dependent on all the input features in the initial fully connected setting. (middle) The learned \ngr optimizes the network connections to fit the regression on the input data as well as satisfy the sparsity constraints, refer Eq.~\ref{eqn:optimization-function-ngr}. If there is a path from the input feature to an output feature, that indicates a dependency, potentially non-linear, between them. The bigger the size of NN (number of layers, hidden unit dimensions) the richer will be the functional representation. Note that not all the weights of the MLP (those dropped during training in grey-dashed lines) are shown for the sake of clarity. (right) The sparse dependency graph between the input and output of the MLP reduces to its normalized weight matrix product $S_{G} = \operatorname{norm}\left(|W_1|\times |W_2|\right)$. }
\label{fig:ngr-architecture}
\vspace{-1mm}
\end{figure}

We choose a multilayer perceptron (MLP) as our neural network choice. The architecture of \ngr is a MLP that takes in input features and fits a regression to get the same features as an output, shown in Fig.~\ref{fig:ngr-architecture}(left). We view this neural network as a glass-box and if there is a path to an output unit (or neuron) from a set of input units, then we interpret that the output unit is a function of those input units. 

In order to obtain a graph, for every feature $X_d$, we want to find the most relevant features that have a direct functional influence on $X_d$. This task becomes increasingly complex as we need to evaluate all possible combinations which can be computationally tedious. Fitting of regression of \texttt{NGRs}, refer Fig.~\ref{fig:ngr-architecture}(middle), can be seen as doing \textit{multitask learning} that simultaneously optimizes for the functional dependencies of all the features.  

Main design challenges or constraints that should be considered while fitting the \ngr regression are:\\ (A) How do we avoid modeling direct self-dependencies among features, eg. $X_d\rightarrow X_d,\forall d\in \{D\}$? \\(B) How do we efficiently induce sparsity among the paths defined by the MLP?


\subsection{Optimization}\label{sec:ngr-opt}

We denote a NN with $L$ number of layers with the weights $\mathcal{W}=\{W_1, W_2, \cdots, W_L\}$ and biases $\mathcal{B}=\{b_1, b_2, \cdots, b_L\}$ as $f_{\mathcal{W, B}}(\cdot)$ with non-linearity not mentioned explicitly. In our implementations, we experimented with multiple non-linearities and found that $\operatorname{ReLU}$ fits well with our framework. Applying the NN to the input $X_\mathcal{D}$ evaluates the following mathematical expression, $f_{\mathcal{W, B}}(X_\mathcal{D}) =\operatorname{ReLU}(W_L\cdot(\cdots(W_2\cdot\operatorname{ReLU}(W_1\cdot X_\mathcal{D} + b_1) + b_2)\cdots)+b_L) 
$.
The dimensions of the weights and biases are chosen such that the neural network input and output units are equal to $\mathcal{D}$ while the hidden layers dimension $H$ remains a design choice. In our experiments, we found a good initial choice of $H=2|\mathcal{D}|$, then eventually based on the loss on validation data, one can adjust the dimensions. 

Our task is to design the \ngr objective function such that it can jointly discover the feature dependency graph constraints (A) \& (B) along with fitting the regression on the input data. We observe that the product of the weights of the neural networks $S_{nn} = { \prod_{l=1}^L }\abs{W_l}=|W_1|\times |W_2|\times \cdots \times |W_L|$ gives us path dependencies between the input and the output units. We note that if $S_{nn}[x_i,x_o]=0$ then the output unit $x_o$ does not depend on input unit $x_i$. 

\textbf{Graph-constrained path norm.} We introduce a way to map NN paths to a predefined graph structure. Consider the matrix $S_{nn}$ that maps the paths from the input units to the output units as described above. Say, we are given a graph with adjacency matrix $S_g\in\{0, 1\}^{D\times D}$. The graph-constrained path norm is defined as $\mathcal{P}_c=\norm{S_{nn}*S_g^c}_1$, where $S_g^c$ is the complement of the adjacency matrix $S_g^c=J_D - S_g$ with $J_D\in\{1\}^{D\times D}$ being an all-ones matrix. The operation $Q*V$ represents the hadamard operator which does an element-wise matrix multiplication between the same dimension matrices $Q$ \& $V$. This term can be used as a penalty term during optimization to fit a particular predefined graph structure, $S_g$.

We utilize the above formulations of MLPs to model the constraints along with finding the set of parameters $\{\mathcal{W, B}\}$ that minimize the regression loss expressed as the Euclidean distance between $X_\mathcal{D}$ to $f_{\mathcal{W, B}}(X_\mathcal{D})$.
The optimization objective becomes 
\begin{align}\label{eqn:ngr-learning}
    \argmin_{\mathcal{W, B}} \sum_{k=1}^{M} \norm{X_{\mathcal{D}}^k - f_{\mathcal{W, B}}(X_\mathcal{D}^k)}^2_2 
    , ~~~~~~ s.t.~~ \operatorname{sym}(S_{nn}) * S_{\text{diag}} = 0
\end{align}
where, $\operatorname{sym}(S_{nn})=\left(\norm{S_{nn}}_2 + \norm{S_{nn}}_2^T\right)/2$ converts the path norm obtained by the NN weights product, $S_{nn}={\prod_{l=1}^L }\abs{W_l}$, into a symmetric adjacency matrix and
$S_{\text{diag}}\in\mathbb{R}^{D\times D}$ represents a matrix of zeroes except the diagonal entries that are $1$. Constraint (A) is thus included as the constraint term in Eq.~\ref{eqn:ngr-learning}. To satisfy the constraint (B), we include an $\ell_1$ norm term $\norm{\operatorname{sym}(S_{nn})}_1$ which will introduce sparsity in the path norms.
Note that this second constraint enforces sparcity of $paths$, not individual $weights$, thus affecting the entire network structure.

We model these constraints as  lagrangian terms which are scaled by a $\log$ function. The $\log$ scaling is done for computational reasons as sometimes the values of the Lagrangian terms can go very low. The constants $\lambda, \gamma$ act as a tradeoff between fitting the regression term and their corresponding constraints. Thus, in order to recover a valid graph dependency structure, the optimization formulation becomes

\begin{equation}\label{eqn:optimization-function-ngr}
    \argmin_{\mathcal{W, B}} \sum_{k=1}^{M} \norm{X_{\mathcal{D}}^k - f_{\mathcal{W, B}}(X_\mathcal{D}^k)}^2_2 + \lambda \norm{\operatorname{sym}(S_{nn}) * S_{\text{diag}}}_1  +\gamma \norm{\operatorname{sym}(S_{nn})}_1
 \end{equation}
 

where, we can optionally add $\log$ scaling to the structure constraint terms. Essentially, we start with a fully connected graph and then the Lagrangian terms induce sparsity in the graph.
Alg.~\ref{algo:NGRs-learning} describes the procedure to learn the \ngr architecture based on optimizing the Eq.~\ref{eqn:optimization-function-ngr}. We note that the optimization and the graph recovered depend on the choices of the penalty constants $\lambda, \gamma$. Since our loss function contains multiple terms, the loss-balancing technique introduced in~\cite{rajbhandari2019antman}, can be utilized to get a good initial value of the constants. Then, while running optimization, based on the regression loss value on a held-out validation data, the values of penalty constants can be appropriately chosen.

\begin{wrapfigure}[17]{R}{0.49\textwidth}
\vspace{-4mm}
\begin{algorithm}[H]
  \DontPrintSemicolon
  \SetKwProg{Fn}{Function}{:}{}
  \SetKwFor{uFor}{For}{do}{}
  \SetKwFor{ForPar}{For all}{do in parallel}{}
  \SetKwFunction{fitNGM}{fit-NGR}
  \SetKwFunction{NGMlearn}{NGR-training}
  \SetKwFunction{proxinit}{proximal-init}
    \Fn{\NGMlearn{$X$}}{
        $f_{\mathcal{W}^0} \gets$ Fully connected MLP \;
        \uFor{$e = 1,\cdots, E$}{
        \text{Xb} $\gets X$  (sample a batch)\;
        $\mathcal{L} = \sum_{k=1}^{M} \norm{\text{Xb}_{\mathcal{D}}^k - f_{\mathcal{W, B}}(\text{Xb}_\mathcal{D}^k)}^2$ \;
        ~~~~~~$- \lambda \norm{\operatorname{sym}(S_{nn}) * S_{\text{diag}}}_1$ \;
        ~~~~~~$- \gamma  \norm{\operatorname{sym}(S_{nn})}_1$\;
        $\mathcal{W}^{e}, \mathcal{B}^e \gets $ backprop $\mathcal{L}$ \& update using `adam' optimizer\;
        }
        $\{\mathcal{W}^E\}\gets f_{\mathcal{W, B}}^E$\;
        $\mathcal{G} \gets \operatorname{sym}\left(\prod_{l=1}^L\abs{W_l^E}\right)$\;
        \KwRet $\mathcal{G}, f_{\mathcal{W}^E}$
    }\vspace{-1mm}
\caption{Learning \ngrs}\label{algo:NGRs-learning}
\end{algorithm}
\end{wrapfigure}



\subsection{Modeling multi-modal data}

It is often common to encounter multi-modal data. For instance, ICU patient records can have information about body vitals (numerical, categorical), nurse notes (natural language) and maybe associated X-rays (images). It will be extremely helpful to get the underlying graph that shows dependencies between these various features. In this section, we propose two different ways to include multi-modal input data in the \ngr formulation. 

\begin{figure}
\centering 
\includegraphics[width=120mm]{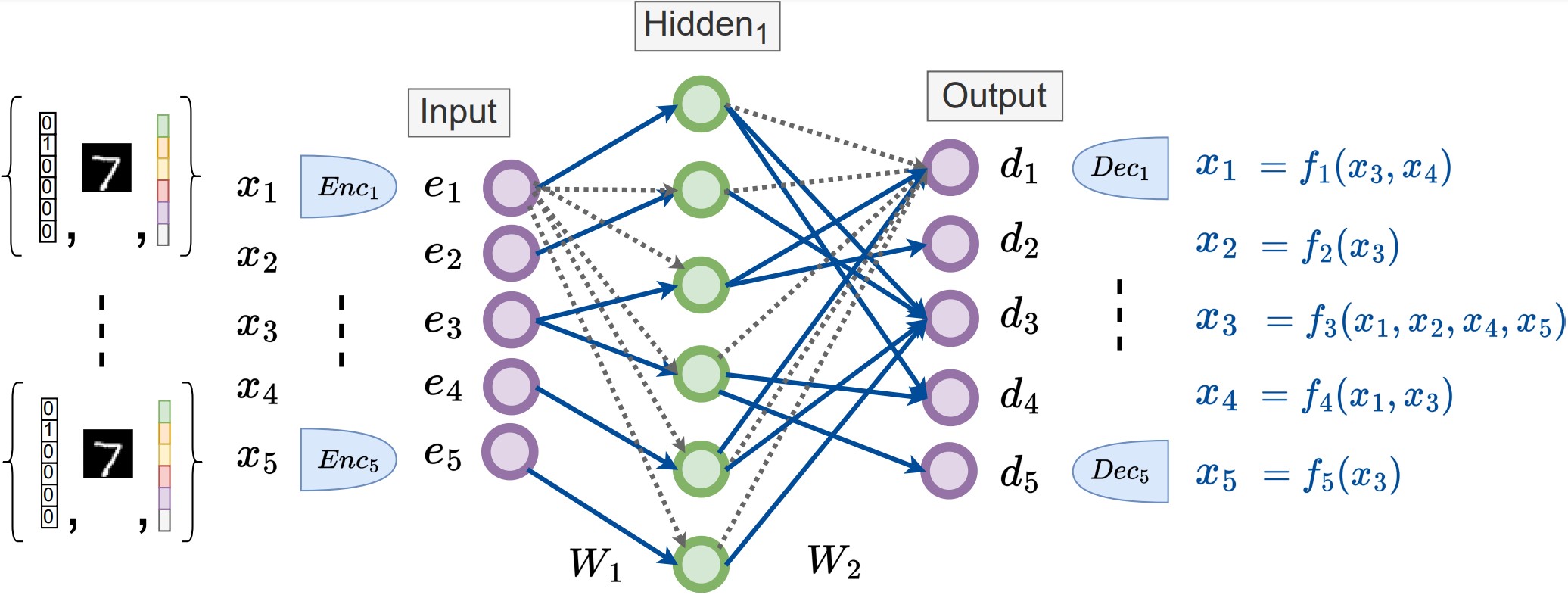}
\caption{\small \textit{Multi-modal data handling with Projection modules.} The input \textbf{X} can be one-hot (categorical), image or in general an embedding (text, audio, speech and other data types). Projection modules (encoder + decoder) are used as a wrapper around the \ngr base architecture. The architecture choice of the projection modules depends on the input data type and users' design choices. Note that the output of the encoder can be more than 1 unit ($e_1$ can be a hypernode) and the corresponding adjacency matrix $S_{\text{diag}}$ of the graph-constrained path norm can be adjusted. Similarly, the decoder side decoder side of the \ngr architecture is updated. The remaining details are similar to the ones described in Fig.~\ref{fig:ngr-architecture}}
\label{fig:ngr-projection}
\end{figure}

\textit{(I) Using projection modules.} Fig.~\ref{fig:ngr-projection} gives a schematic view of including projection modules to the base architecture described in Fig.~\ref{fig:ngr-architecture}. W.l.o.g. we can consider that each of the $D$ inputs is an embedding in $x_i\in\mathbb{R}^I$ space. For example, given an image, one way of getting a corresponding embedding can be to use the latent layer of a convolutional neural network based autoencoder. We convert all the input $x_i$ nodes in the \ngr architecture to hypernodes, where each hypernode contains the embedding vector. Consider a hypernode that contains an embedding vector of size $\mathbb{R}^E$ and if an edge is connected to the hypernode, then that edge is connected to all the $E$ units of the embedding vector. For each of these input hypernodes, we define a corresponding encoder embedding $e_i\gets\text{enc}_i(x_i),\forall e_i\in\mathbb{R}^E$, which can be designed specific to that particular input embedding. Similarly, we apply the encoder modules to all the $x_i$ hypernodes and obtain the $e_i$ hypernodes. Same procedure is followed at the decoder end, where $x_i\gets\text{dec}_i(d_i), \forall d_i\in\mathbb{R}^O$. We do this step, primarily to reduce the input dimensions. Now, the \ngr graph discovery optimization reduces to discovering the connectivity pattern using the path norms between hypernodes $e_i$'s and $d_i$'s. A slight modification to the graph-constrained path norm is needed to account for the hypernodes. The $S_{diag}$ term will now represent the connections between the hypernodes, so $S_{diag}\in\{0, 1\}^{DE\times DO}$ with ones in the block diagonals. 
We can include the projection modules in the regression term of the \ngr objective function, while the structure learning terms will remain intact
\begin{equation}\label{eqn:proj-optimization-function-ngr}
    \argmin_{\mathcal{W, B, \operatorname{proj}}} \sum_{k=1}^{M} \norm{X_{\mathcal{D}}^k - f_{\mathcal{W, B,\operatorname{proj}}}(X_\mathcal{D}^k)}^2_2 + \lambda \norm{\operatorname{sym}(S_{nn}) * S_{\text{diag}}}_1 + \gamma \norm{\operatorname{sym}(S_{nn})}_1
 \end{equation}
where the $\operatorname{proj}$ are the parameters of the encoder and decoder projection modules as discussed in Fig.~\ref{fig:ngr-projection}.

\begin{figure}
\centering 
\includegraphics[width=120mm]{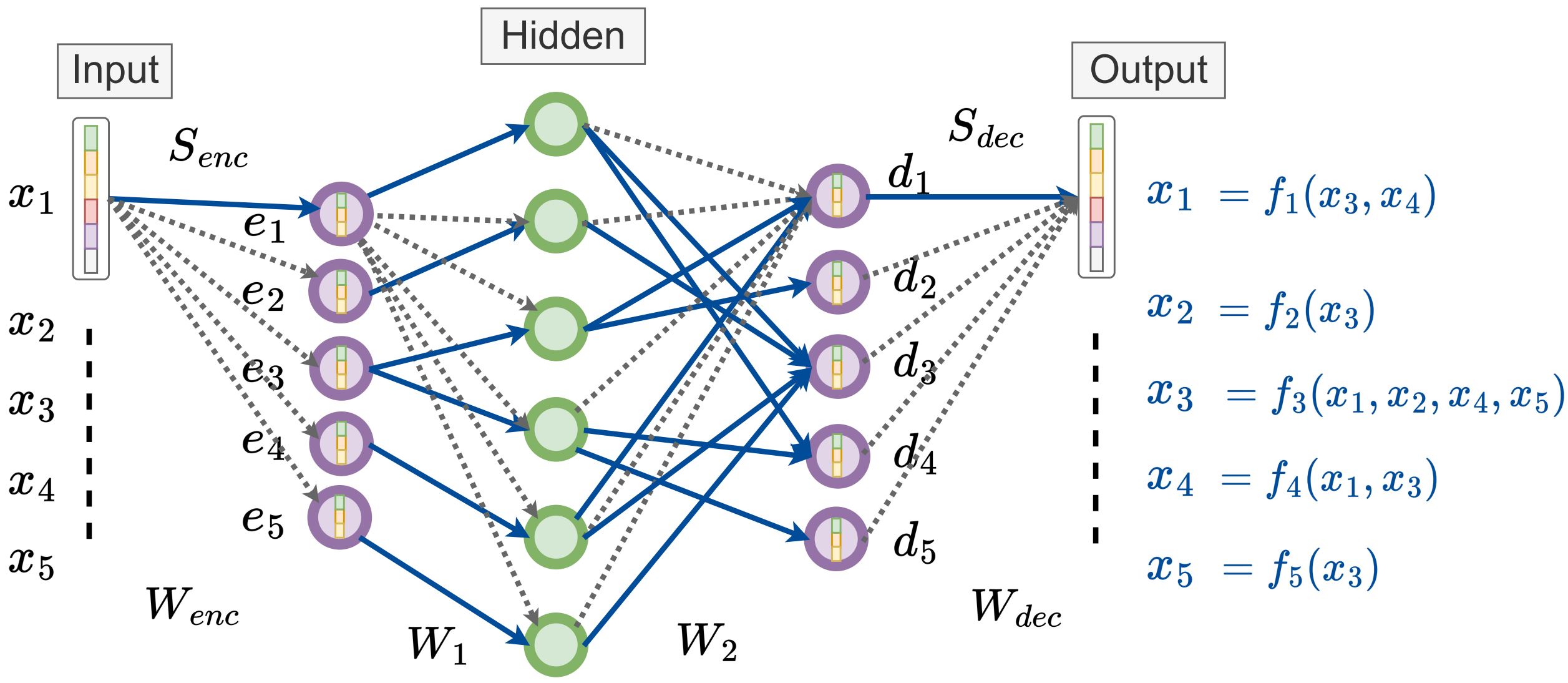}
\caption{\small \textit{Multi-modal data handling with Graph-constrained path norm.} W.l.o.g. we consider an input \textbf{X} to be embeddings that can come from text, audio, speech and other data types. We extend the idea of applying \gcpn to the encoder MLP and the decoder MLP. We initilialize a fully connected MLP and then using the \gcpn penalties, we capture the desired input to output unit path dependencies after optimizing the Eq.~\ref{eqn:gcpn-optimization-function-ngr}. Neural network nodes containing embeddings are shown as hypernodes. We define hypernodes for the sake of brevity to convey that all units of the embedding vector within the hypernode are considered a single unit when deciding the edge connections defining a graph. The encoder and decoder MLPs are used as a wrapper around the \ngr base architecture. The remaining details are similar to the ones described in Fig.~\ref{fig:ngr-architecture}.}
\label{fig:ngr-gcpn}
\end{figure}

\textit{(II) Using Graph-constrained path norm.} Fig.~\ref{fig:ngr-gcpn} shows that we can view the connections between the $D$ hypernodes of the input embedding $x_i\in\mathbb{R}^I$ to the corresponding input of the encoder layer $e_i\in\mathbb{R}^E$ as a graph. We represent these set of input layer to the encoder layer connections by $S_{\text{enc}}\in\{0, 1\}^{DI\times DE}$, where there is a $S_{\text{enc}}[x_i, e_j]=1$ if the $(x_i, e_j)$ hypernodes are connected. So, if we initialize a fully connected neural network (or MLP) between the input layer and the encoder layer, we can utilize the \gcpn penalty function to map the paths of the input units to the encoder units to satisfy the graph structure defined by $S_{\text{enc}}$. Similar exercise is replicated at the decoder end to obtain $S_{\text{dec}}$. This extension of the \gcpn to multi-modal data leads us to the following Lagrangian based formulation of the optimization objective
\begin{align}\label{eqn:gcpn-optimization-function-ngr}
    &\argmin_{\mathcal{W_\text{enc}, W, B, W_\text{dec}}} \sum_{k=1}^{M} \norm{X_{\mathcal{D}}^k - f_{\mathcal{W_\text{enc}, W, B, W_\text{dec}}}(X_\mathcal{D}^k)}^2_2 + \lambda \norm{\operatorname{sym}(S_{nn}) * S_{\text{diag}}}_1 \\
    &\qquad+ \gamma \norm{\operatorname{sym}(S_{nn})}_1 
    + \eta\norm{\operatorname{sym}(S_{nn}^e) * S_{\text{enc}}}_1 + \beta \norm{\operatorname{sym}(S_{nn}^d) * S_{\text{dec}}}_1\nonumber
\end{align}
where $f_{\mathcal{W_\text{enc}, W, B, W_\text{dec}}}(\cdot)$ represents the entire end-to-end MLP including the encoder and decoder mappings, $S_{nn}^e={ \prod_{l=1}^{L^e} }\abs{W_l}=|W_1|\times |W_2|\times \cdots \times |W_{L^e}|$ captures the path dependencies in the encoder MLP with $L^e$ layers, $S_{nn}^d={ \prod_{l=1}^{L^d} }\abs{W_l}=|W_1|\times |W_2|\times \cdots \times |W_{L^d}|$ captures the path dependencies in the decoder MLP with $L^d$ layers. The Lagrangian constants $\lambda, \gamma, \eta, \beta$ are initialized in the same manner as outlined in Sec.~\ref{sec:ngr-opt}. We note the advantage of using the \gcpn penalties to enable \textbf{soft enforcing} of the path constraint requirements between the input and output units of the neural networks. We recommend the \gcpn based approach (II) as the implementation is straightforward, is highly scalable and can handle large embedding sizes. 



\subsection{Representation as a probabilistic graphical model}

Once the sparse graph is recovered, the learned architecture of \ngrs represent the functional dependencies between the features. A beneficial next step for wider adoption will be the ability to model the entire joint probability distribution of the features. This type of representation of the functional dependencies based on neural networks has been recently explored in~\cite{shrivastava2022neural} and are termed as `Neural Graphical Models' (\texttt{NGMs}). They are a type of probabilistic graphical models that utilize neural networks and a pre-defined graph structure between features to learn complex non-linear underlying distributions. Additionally, they can model multi-modal data and have efficient inference \& sampling algorithms. The inference capability can be used to estimate missing values in data. The learned \ngr model can be viewed as an instance of \ngmsns. 


\section{Experiments}

\subsection{Learning Gaussian Graphical Models}

\begin{figure}
\centering 
\includegraphics[width=0.2\textwidth, height=40mm]
{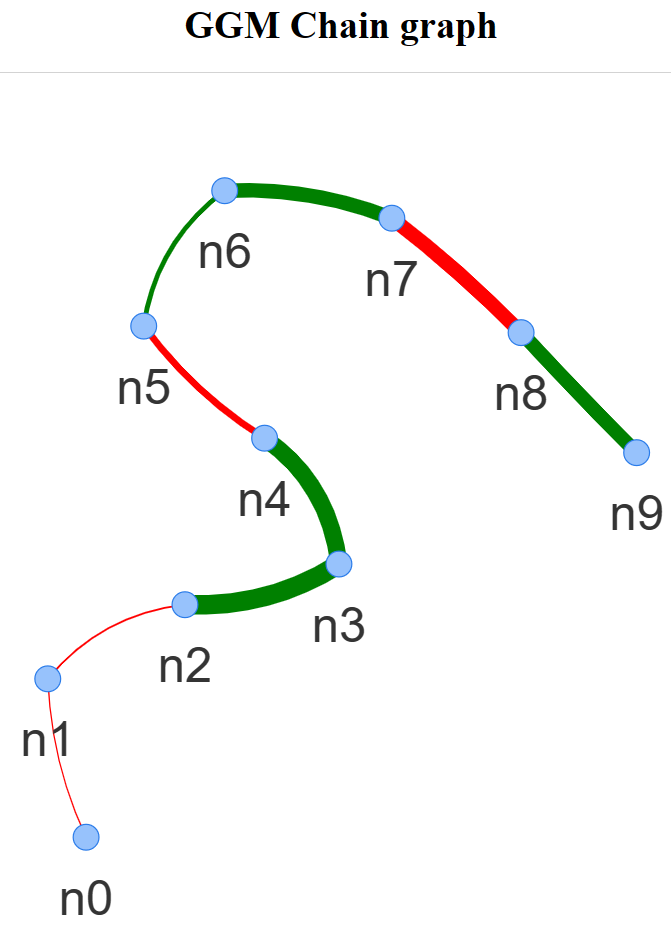}~
\includegraphics[width=0.4\textwidth,height=40mm]{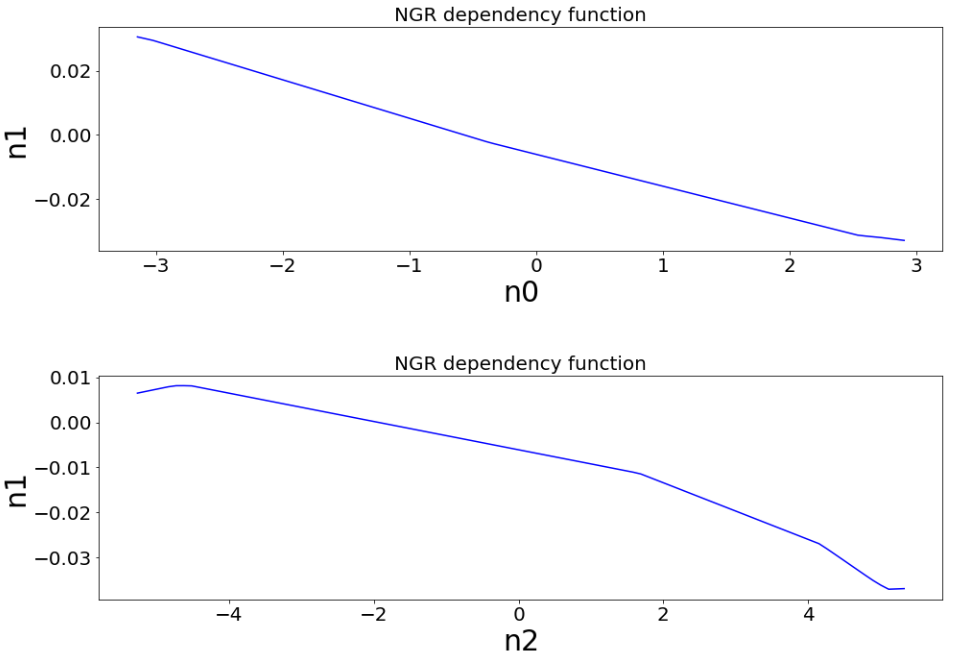}~
\includegraphics[width=0.4\textwidth,height=40mm]{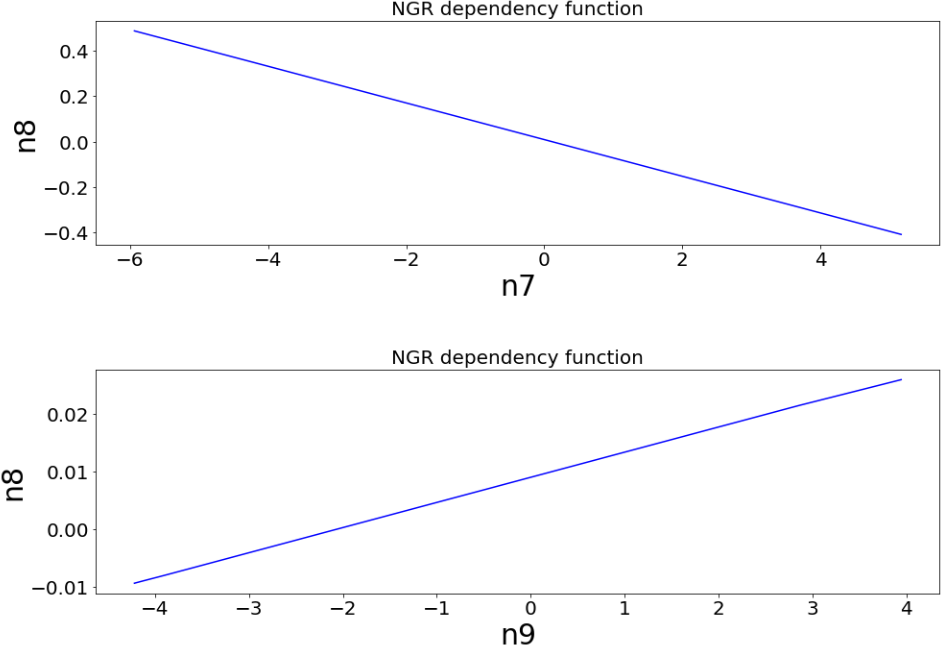}
\caption{\small \textit{Modeling GGMs using \ngrsns.} (left) The Conditional Independence graph~\cite{shrivastava2022methods} for the chain structure is shown. Positive partial correlations between the nodes are shown in green, while the negative partial correlations in red. A positive partial correlation between nodes (A, B) will mean that increasing the value of A will correspond to increase in value of B. Partial negative correlation will mean the opposite. These correlations show direct dependence or, in other words, the dependence is evaluated conditioned on all the other nodes. (middle, right) We observe that the NGR slopes match the trend in the GGM graph. This shows that the dependency plots learned comply with the desired behaviour as shown in the color of the partial correlation edges.}
\label{fig:ngr-ggm}
\end{figure}

We explore the \ngrns's ability to model Gaussian Graphical Models (GGM). To create a GGM, we used a chain-graph structure and then defined a precision matrix over it by randomly initializing the entries $~\mathcal{U}(0.5, 1)$ with random signs. The diagonal entries of the precision matrix were chosen, such that it is Positive Semi-definite. Samples were obtained from the GGM and were used as input to the \ngr for recovering the underlying graphical model.  Fig.~\ref{fig:ngr-ggm} shows the GGM and the corresponding trends discovered after fitting a \ngrns. We used a \ngr with a single hidden layer and its dimension $H=100$. Table~\ref{tab:ngr-ggm-expt} show the graph recovery results by running \ngr on varying number of samples obtained using the Gaussian graphical model. As expected, the results improve as we increase the number of samples and thereby \ngrs are capable of representing Gaussian graphical models.

\textit{Note on training.} We found that learning of \ngrs can be sensitive to the hyperparameter choices like learning rate, optimizer choice (we primarily used `adam') and the hidden layer dimension size. The choice of $\lambda$, which balances between the regression loss and the structure losses is tricky to decide. When optimizing \ngrns, we observe that it first optimizes the regression loss, then when the regression loss stabilizes, the structure loss gets optimized which increases the regression loss a bit and then eventually both the losses go down smoothly. So, our general approach to train \ngrs is to choose large number of training epochs with low learning rates.  

\begin{table}[h]
\floatbox[{\capbeside\thisfloatsetup{capbesideposition={left, center},capbesidewidth=8.5cm}}]{table}[0.999\FBwidth]
{\caption{\small The recovered CI graph from \ngr is compared with the ground truth CI graph defined by the underlying GGMs precision matrix with $D=10$ nodes, chain graph as shown in Fig.~\ref{fig:ngr-ggm}. Area under the ROC curve (AUC) and Area under the precision-recall curve (AUPR) values for 5 runs are reported.}\label{tab:ngr-ggm-expt}}
{
\centering
\resizebox{0.35\textwidth}{!}{
\centering
\begin{tabular}{|c|c|c|}
\hline
Samples & AUPR & AUC \\ \hline
100 & $0.34\pm0.03$ & $0.67\pm 0.05$ \\ \cline{1-1}
500 & $0.45\pm 0.10$ & $0.79\pm0.03$  \\ \cline{1-1}
1000 & $0.63\pm0.11$ & $0.90\pm 0.03$ \\ \hline
\end{tabular}
}
}
\end{table}

\subsection{Infant Mortality data analysis}\label{sec:infant-mortality}

The infant mortality dataset we used is based on CDC Birth Cohort Linked Birth – Infant Death Data Files \cite{CDC:InfantLinkedDatasets}. It describes pregnancy and birth variables for all live births in the U.S. together with an indication of an infant's death (and its cause) before the first birthday.  We used the data for 2015 (latest available), which includes information about 3,988,733 live births in the US during 2015 calendar year. We recoded various causes of death into 12 classes: alive, top 10 distinct causes of death, and death due to other causes. 

\begin{figure}
\centering 
\includegraphics[width=45mm]{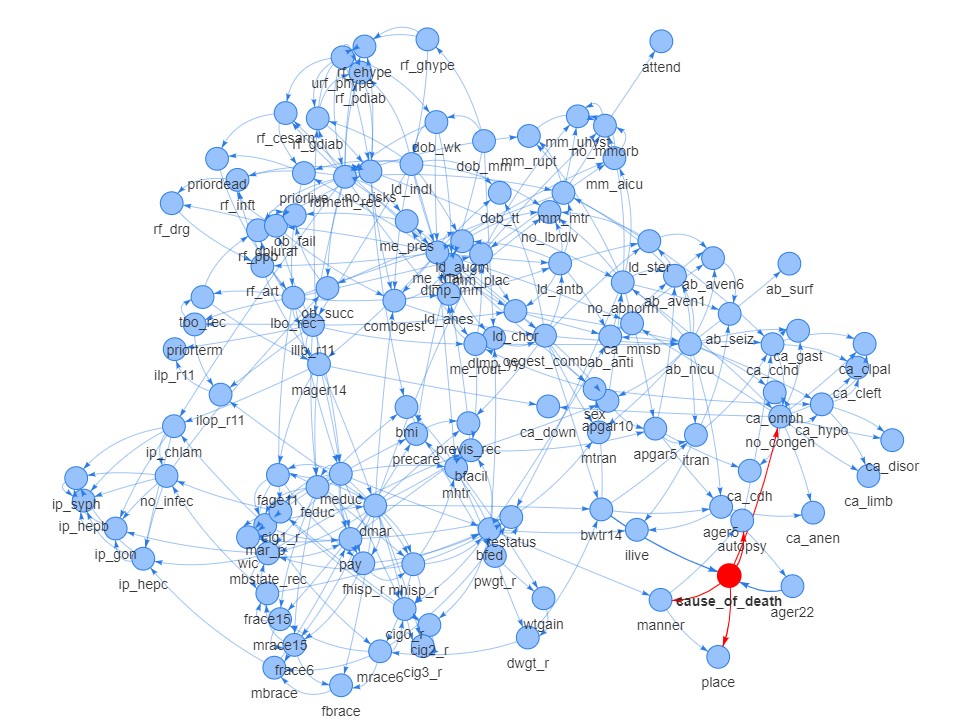}
\includegraphics[width=45mm]{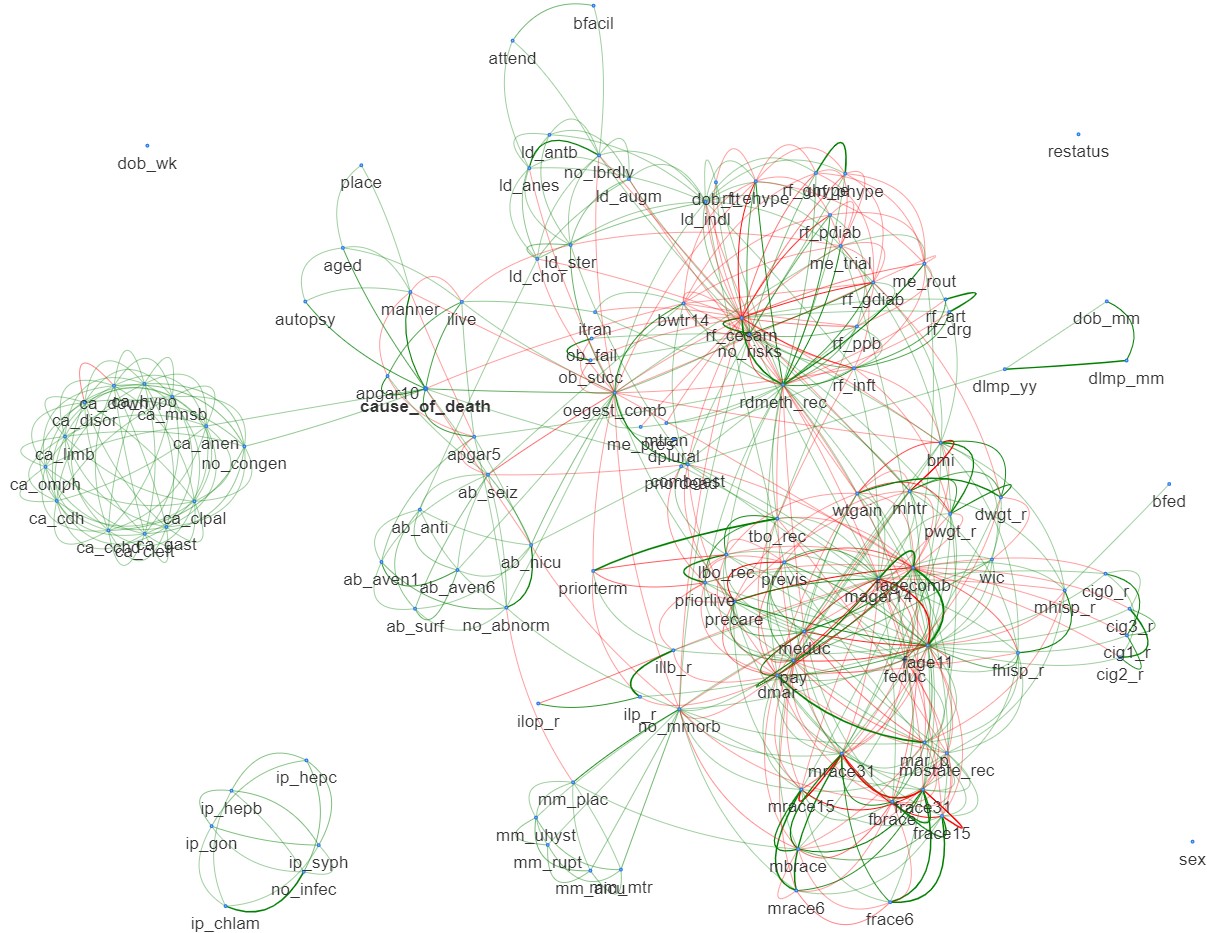}
\includegraphics[width=45mm]{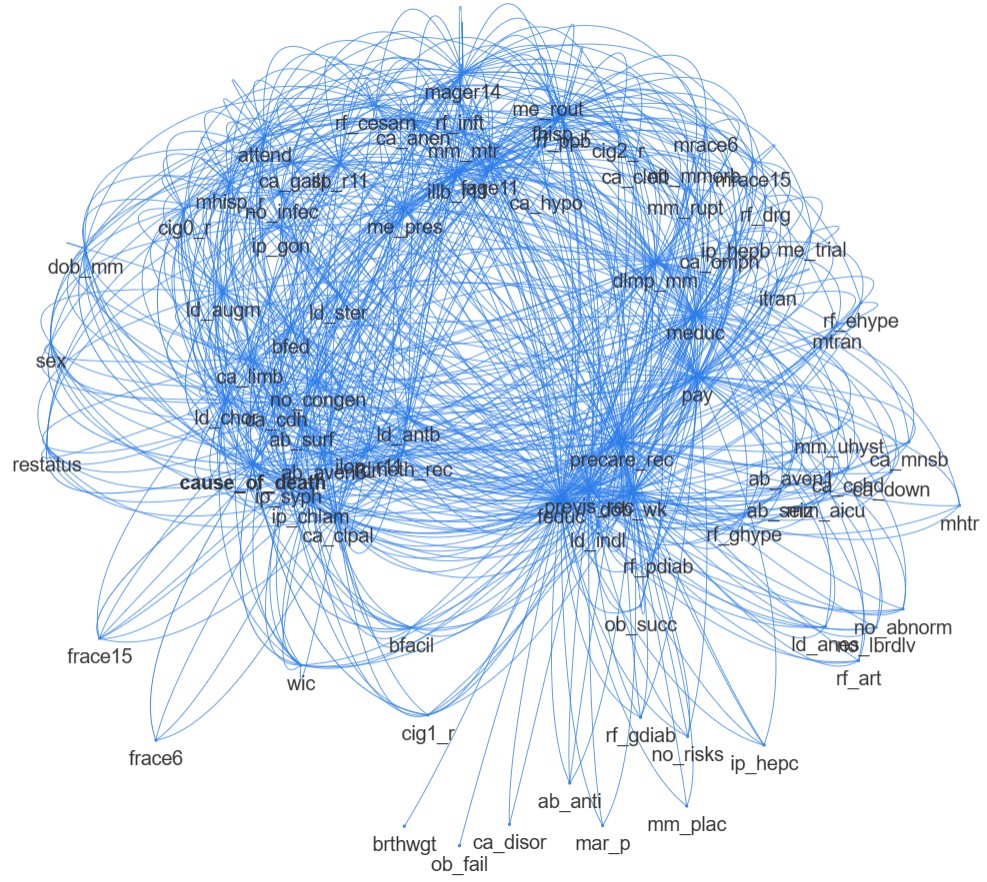}
\caption{\small \textit{Graphs recovered for the Infant Mortality 2015 data.} (left) The Bayesian network graph learned using score-based method, (middle) the CI graph recovered by \uglad and (right) the \ngr graph. For \ngrns, we applied a threshold to retain top relevant edges. }
\label{fig:im2015-graphs}
\end{figure}

\textit{Recovered graphs}.
We recovered the graph strucure of the dataset using \ngrns, \uglad \citep{shrivastava2022uglad} and using Bayesian network package \texttt{bnlearn}~\citep{bnlearn} with Tabu search and AIC score.  The graphs are shown in Fig.~\ref{fig:im2015-graphs}. All variables were converted to categorical for \texttt{bnlearn} structure learning and inference as it does not support networks containing both continuous and discrete variables. In contrast, \uglad and \ngrs are both equipped to work with mixed types of variables and were trained on the dataset prior to conversion. It is interesting to observe that although there are some common clusters in all three graphs (parents' race and ethnicity (\texttt{mrace} \& \texttt{frace}); variables related to mother's bmi, height (\texttt{mhtr}) and weight, both pre-pregnancy (\texttt{pwgt\_r}) and at delivery (\texttt{dwgt\_r}); delivery route (\texttt{rdmeth\_rec}); type of insurance (\texttt{pay}); parents' ages (\texttt{fage} and \texttt{mage} variables); birth order (\texttt{tbo} and \texttt{lbo}); prenatal care etc.), each graph has a different extent of inter-cluster connections. Since, all the three different graph recovery methods are based on different distribution assumptions, training methodology, way of handling multimodal data leads to getting different connectivity patterns. It becomes increasingly important to have efficient inference algorithms for these models to do probabilistic queries in order to utilize them effectively and extract insights. This dataset and corresponding BN, \uglad graphs analysis are discussed in~\cite{shrivastava2022neural}.

\begin{table*}[]
\centering
\vspace{-4mm}
\caption{\small Comparison of predictive accuracy for gestational age and birthweight.} 
\label{tab:inference_results1}
\resizebox{0.7\textwidth}{!}{
\begin{tabular}{|c|c|c|c|c|}
\hline
Methods & \multicolumn{2}{c|}{Gestational age} & \multicolumn{2}{c|}{Birthweight}  \\ 
& \multicolumn{2}{c|}{(ordinal, weeks)} & \multicolumn{2}{c|}{(continuous, grams)}  \\ \hline
& MAE & RMSE & MAE & RMSE \\ \hline
Logistic Regression & $1.512\pm0.005$ & $3.295\pm0.043$ & N/A & N/A  \\ \hline
Bayesian network & {\boldmath $1.040 \pm0.003$} & $2.656\pm0.027$ & N/A & N/A \\ \hline
EBM & $1.313 \pm0.002$ & $2.376\pm0.021$ & { \boldmath $345.21\pm1.47$ } & {\boldmath $451.59\pm2.38$}  \\ \hline
\ngm w/BN graph & $1.364 \pm0.025$ & $2.452\pm0.026$ & $370.20\pm1.44$ & $484.82\pm1.88$  \\ \hline
\ngm w/\uglad graph & $1.295 \pm0.010$ & {\boldmath $2.370\pm0.025$ } & $371.27\pm1.78$ & $485.39\pm1.86$  \\ \hline
\ngr & $1.448 \pm0.133$ & $2.493\pm0.100$ & $369.68\pm1.14$ & $483.96\pm1.56$  \\ \hline
\end{tabular}}
\label{tab:inference_results}
\vspace{-2mm}
\end{table*}

\begin{table*}[]
\centering
\caption{\small Comparison of predictive accuracy for 1-year survival and cause of death.  Note: recall set to zero when there are no labels of a given class, and precision set to zero when there are no predictions of a given class.} 
\resizebox{\textwidth}{!}{
\begin{tabular}{|c|c|c|c|c|c|c|}
\hline
Methods & \multicolumn{2}{c|}{Survival} & \multicolumn{4}{c|}{Cause of death} \\ 
& \multicolumn{2}{c|}{(binary)} & \multicolumn{4}{c|}{(multivalued, majority class frequency $0.9948$)} \\ \hline
& \multicolumn{2}{c|}{} & \multicolumn{2}{c|}{micro-averaged}  & \multicolumn{2}{c|}{macro-averaged}  \\ 
& AUC & AUPR & Precision & Recall & Precision & Recall\\ \hline
Logistic Regression & $0.633\pm0.004$ & $0.182\pm0.008$ & $0.995\pm7.102\text{e-}05$ & $0.995\pm7.102\text{e-}05$ & $0.136\pm0.011$ & $0.130\pm0.002$ \\ \hline
Bayesian network & $0.655\pm0.004$ & $0.252\pm0.007$ & $0.995\pm7.370\text{e-}05$  & $0.995\pm7.370\text{e-}05$ &  $0.191\pm0.008$ & $0.158\pm0.002$\\ \hline
EBM & $0.680\pm0.003$ & {\boldmath $0.299\pm0.007$ } & $0.995\pm5.371\text{e-}05$ &  $0.995\pm5.371\text{e-}05$ &  $0.228\pm0.014$ & $0.166\pm0.002$ \\ \hline
\ngm w/BN graph & $0.752\pm0.012$ & $0.295\pm0.010$ & $0.995\pm4.416\text{e-}05$ & $0.995\pm4.416\text{e-}05$ & $0.497\pm2.208\text{e-}05$ & {\boldmath $0.500\pm1.000\text{e-}06$ }\\ \hline
\ngm w/\uglad graph & $0.726\pm0.020$ & $0.269\pm0.018$ & $0.995\pm9.735\text{e-}05$ & $0.995\pm9.735\text{e-}05$ & $0.497\pm4.868\text{e-}05$ & {\boldmath $0.500\pm1.000\text{e-}06$} \\ \hline
\ngr & {\boldmath $0.770\pm0.009$} & $0.269\pm0.030$ & {\boldmath $0.995\pm3.357\text{e-}05$} & {\boldmath $0.995\pm3.357\text{e-}05$} & {\boldmath $0.497\pm1.678\text{e-}05$} & {\boldmath $0.500\pm1.000\text{e-}06$} \\ \hline
\end{tabular}}
\label{tab:inference_results2}
\vspace{-2mm}
\end{table*}

\textit{\ngr architecture details.} Since we have mixed input data types, real and categorical data, we utilize the \ngr multimodal architecture's neural view given in Fig.~\ref{fig:ngr-gcpn}. We used a 2-layer neural view with $H=1000$. The categorical input was converted to its one-hot vector representation and added to the real features which gave us roughly $\sim 500$ features as input. The neural view input from the encoder had the same dimension as input. Same dimension settings were replicated at the decoder end. \ngr was trained on the 4 million data points with ~$D=500$ using 64 CPUs within 4 hours.  

\textit{Inference accuracy comparison}. Infant mortality dataset is particularly challenging due to the data skewness. For instance, the cases of infant death during the first year of life are rare compared to cases of surviving infants. Getting good performance on imbalanced data is a challenging problem and multiple techniques have been developed to assist existing learning algorithms~\cite{chawla2002smote,shrivastava2015classification,bhattacharya2017icu}. In our case, we directly apply the models as is for obtaining results on the base implementation. Since \ngr becomes an instance of a neural graphical model, we also include comparisons of  \ngms that use base graphs obtained from Bayesian Networks and CI graph from \ugladns. We compared prediction for four variables of various types:  gestational age (ordinal, expressed in weeks), birthweight (continuous, specified in grams), survival till 1st birthday (binary) and cause of death ('alive', 10 most common causes of death with less common grouped in category 'other' with 'alive' indicated for 99.48\% of infants). We compared with other prediction methods like  logistic regression, Bayesian networks, Explainable Boosting Machines (EBM)~\cite{caruana2015intelligible, lou2013accurate} and report 5-fold cross validation results. 

Tables~\ref{tab:inference_results1} and~\ref{tab:inference_results2} demonstrate that \ngr models are more accurate than logistic regression, Bayesian Networks and on par with EBM models for categorical and ordinal variables. They performance is at par to the \ngm models with different input base graphs highlighting that learning a \ngr graph is can help us gain new insights. We note an additional advantage of \ngrs and \ngms in general that we just need to train a single model and their inference capability can be leveraged to output predictions for the tasks listed here. For the case of EBM model, we had to train a separate model for each outcome variable evaluated.

\section{Conclusions, Discussions \& Future work}

We address the important problem of doing sparse graph recovery and querying the corresponding graphical model. The existing graph recovery algorithms make simplifying assumptions about the feature dependency functions in order to achieve sparsity. Some deep learning based algorithms achieve non-linear functional dependencies but their architecture demands too many learnable parameters. After obtaining the underlying graph representation, running probabilistic inference on such models is often unclear and complicated. Neural Graph Revealers leverage neural networks to represent complex distributions and also introduces the concept of Graph-constrained path norm to learn a sparse graphical model. They  are an instance of special type of neural networks based PGMs known as Neural Graphical Models which provide efficient algorithms for downstream reasoning tasks. Our experiments on infant mortality dataset demonstrate usefulness of \ngrs to model complex multimodal input real-world problems. 

\textbf{Effects of class imbalance in data on graph recovery}: In our experiments, we find that the graphs recovered are sensitive to such class imbalance. We have noticed that \textbf{none} of the prior graph recovery methods identify this issue. The \ngr architecture provides flexibility in handling such class imbalanced data. We can leverage the existing data augmentation techniques~\cite{chawla2002smote,fernandez2018smote} or cost function balancing techniques between the different classes data points~\cite{shrivastava2015classification,bhattacharya2017icu,bhattacharya2019methods,shrivastava2021system}. Besides, the sparsity induced in the neural network architecture because of the \gcpn forces the \ngrs to focus on important trends and thus provides a certain extent of robustness to noisy data. It is an interesting future direction that we are actively exploring for \ngrsns.

\textbf{Utilizing Graph connections for Randomized Neural Networks}: The research on sparse graph recovery along with the associated graphical model obtained using Neural Graph Revealers and their corresponding Neural Graphical Models opens up a possibility of finding connections with the theory of randomized neural networks~\cite{frankle2018lottery,shrivastava2021echo,gallicchio2020deep,ramanujan2020s}. Especially, we obtain a sparse MLP with \ngrsns, which can be interesting to draw parallels with the theories like `Lottery Ticket Hypothesis' and randomized neural networks. We are investigating this idea further.

\textbf{Video Processing \& Text Mining} Videos can be interpreted as a sequential collection of frames~\cite{oprea2020review}. Graph based methods can model the feature interactions within a single frame and inter-frame evolution over the sequence of frames~\cite{shrivastava2024methods}. Such methods provide a strong alternative approach for various video related tasks~\cite{jiao2021new,bodla2021hierarchical,saini2022recognizing,zhou2022survey}. \ngrs can be helpful for video prior representation~\cite{shrivastava2024video2,shrivastava2024video3} as it can capture internal characteristics over the video frames. As \ngrns-generic architecture can handle multi-modal input, it can be tweaked to model continuous multi-dimensional processes as described in~\cite{shrivastava2024video1}. Generating diverse video ~\cite{denton2018stochastic,shrivastava2021diverse,shrivastava2021diversethesis} has recently garnered considerable attention where \ngrs can play a key role of estimating the features of the next predicted frame. Sparse graphs can be used in conjunction with text mining tools like~\cite{fize2017geodict,roche2017valorcarn,antons2020application}, where \ngrs can be useful in extracting dependency insights among the text datasets.

\textbf{\ngr design choices}: During the process of designing \ngrsns, we tried multiple design choices to narrow down to the architecture that we present in this paper. We include some of those details here in order to increase transparency and facilitate further research and adoption of \ngrns. 
\begin{itemize}[leftmargin=*,nolistsep]
    \item Various choices of the structure enforcing terms $\norm{\operatorname{sym}(S_{nn}) * S_{\text{diag}}}, \norm{\operatorname{sym}(S_{nn})}$ were tried. This include applying $\ell_1$ norm, Frobenius norm and other higher order norms. We also find the nonzero counts of these matrices to be an interesting choice.
    \item We optionally tried enforcing symmetry of the output $S_{nn}$ or the adjacency matrix by adding a soft threshold term $\norm{S_{nn} - S_{nn}^T}$ with various norms.  
    \item It is tricky to optimize the objective function since it contains multiple terms. To ensure that all the losses are reduced, we recommend using large number of epochs and lower learning rates as a conservative choice. Optionally, we find that log scaling of the structure penalty terms can be useful in some cases. 
\end{itemize}

We believe that this direction of research can lead to a wider adoption of the modeling and storing data in a compact format as the graphical models designs are intended to do. Our approach is a direction towards developing a novel representation of PGMs, improving efficiency of doing probabilistic reasoning, inference and sampling.

\bibliography{citations,bibfile}

\bibliographystyle{iclr2023_conference}

\end{document}